\documentclass[10pt,twocolumn,letterpaper]{article}

\usepackage{cvpr}
\usepackage{times}
\usepackage{epsfig}
\usepackage{graphicx}
\usepackage{amsmath}
\usepackage{amssymb}

\usepackage[pagebackref=true,breaklinks=true,letterpaper=true,colorlinks,bookmarks=false]{hyperref}
\newcommand{\minisection}[1]{\vspace{0.04in} \noindent {\bf #1}\ \ }

\cvprfinalcopy 
\def\cvprPaperID{****} 

\ifcvprfinal\fi

\begin{document}

\title{Soft-PHOC Descriptor for End-to-End Word Spotting \\ in Egocentric Scene Images} 

\author{Dena Bazazian, Dimosthenis Karatzas\\
CVC,
Universitat Aut\`onoma de Barcelona\\
{\{dena.bazazian, dimos\}@cvc.uab.es}
\and
Andrew D. Bagdanov\\
MICC,
University of Florence\\
{andrew.bagdanov@unifi.it}
}
\maketitle

\begin{abstract}
  Word spotting in natural scene images has many applications in scene
  understanding and visual assistance. In this paper we propose
  a technique to create and exploit an intermediate representation of
  images based on text attributes which are character
  probability maps. Our representation extends the concept of the
  Pyramidal Histogram Of Characters (PHOC) by exploiting Fully
  Convolutional Networks to derive a pixel-wise mapping of the
  character distribution within candidate word regions. We call this
  representation the \textit{Soft-PHOC}. Furthermore, we show how to
  use Soft-PHOC descriptors for word spotting tasks in egocentric
  camera streams through an efficient text line proposal
  algorithm. This is based on the Hough Transform over character
  attribute maps followed by scoring using Dynamic Time Warping (DTW).
  We evaluate our results on ICDAR 2015 Challenge 4 dataset of
  incidental scene text captured by an egocentric camera \footnote{Our source code is publicly available under: \url{https://github.com/denabazazian/SoftPHOC_TextDescriptor}.}.
\end{abstract}

\section{Introduction}

Reading text in the wild is an important task in many computer vision
applications as text carries semantically rich information about scene
content and context. For instance, egocentric cameras have been used
for assisting visually impaired people by reading text detected in the
scene~\cite{Ani17}. Furthermore, being able to determine the presence
or absence of given words can improve the understanding of the
surrounding context or provide detailed information about objects in
the scene. Significant advances in the state-of-the-art in scene text
recognition have been made in recent years~\cite{Liu18,Ma18,Shi18}. However, the
bulk of the emphasis has been on the closed dictionary setting which
exploits a large dictionary of known words for training.

In this paper we address the challenge of spotting text in egocentric
scene images. At the same time, we propose a general framework without restricting the recognizable words to a fixed lexicon or
dictionary.

Words which are typically out-of-dictionary include, for
instance, price tags, telephone numbers, URLs, dates or other cases
where punctuation marks are present in the words. To be able to
recognize this kind of structured text, a character based
representation of words is needed, since we can not rely on a
restricted collection of words. The contributions of this paper are:\\
-- we introduce a novel mid-level word representation, we
  call it the Soft-PHOC and this representation captures the intra-word character
  dependencies;\\
-- we propose a training strategy to learn to effectively encode unlabeled images;\\
-- we present a new proposal approach for text detection which is based on text line instead of bounding box, in this case we employ the Hough Transform in lieu of bounding box generators; and\\
-- we propose a novel technique to extract multi oriented
  bounding boxes for text detection based on the text lines and their
  orientation.\\


The robustness of this approach stems from the capacity of the
Soft-PHOC encoder to simultaneously represent each character
individually and the entire word. We argue that detecting a
query word in scene images with a bounding box may be inconsequential,
and that the same level of information can be obtained by localizing
the query with just a line.

The remainder of the article is organized as follows. In the next
section we review work related to our approach. In
Section~\ref{sec:PropMethod} we describe our Soft-PHOC descriptor, and
in Section~\ref{sec:HoughDTW} we show how to use a Hough Transform and
Dynamic Time Warping in lieu of bounding box proposals for word
spotting. We report our experimental results in
Section~\ref{sec:ExpResults}, and finally discuss our
contribution and draw some conclusions in
Section~\ref{sec:Conclusion}.
\section{Related Work}
\label{relatedWork}
Word spotting in scene images recently attracts a lot of attention in document image understanding. 
In this section, we present a brief introduction to
related works including text detection, text recognition and word
spotting methods that combine both.

\minisection{Proposal-based text recognition} 

\noindent
Deep Convolutional Neural Networks (DCNNs) have become the standard approach for many
computer vision tasks, and DCNN methods are also state-of-the-art for
text recognition. The authors of~\cite{Jaderberg16} studied about the
problem of unconstrained text recognition using generic object
proposals and a CNN to recognize words from an extensive
lexicon. However, the generic object proposal approach does not
perform well on text detection tasks. The Text Proposals
approach~\cite{Gomez16} introduced a text-specific object proposal
method that is based on generating a hierarchy of word hypotheses
according to the similarity region grouping algorithm. Later, the
authors of~\cite{Bazazian16} fused the Text Proposals technique with a
Fully Convolutional Network (FCN)~\cite{Shelhamer16} in order to
achieve high text region recall while considering significantly fewer
candidate regions. In a follow-up work~\cite{Bazazian17} they improved
the pipeline to increase the speed of the text proposal
generator. They also demonstrated the optimal performance of the text
detector in comparison with the state-of-the-art general object detector
technique~\cite{Redmon17} on text detection tasks.  This approach has
been applied to compressed images~\cite{Galteri17}.
TextBoxes~\cite{Liao17} re-purposed the SSD detector~\cite{Liu16} for
word-wise text localization. Exploiting the robustness SSD, the
authors of~\cite{He17} proposed an attention mechanism that directly
detects the word-level bounding box. Gupta et al. in~\cite{Gupta16} generated synthetic data and propose an architecture inspired by FCN and YOLO~\cite{Redmon16}. Ma et al. in~\cite{Ma18} adapt the Faster R-CNN architecture and extend it to detect text of different orientations by adding anchor boxes of 6 hand-crafted rotations and 3 aspect ratios. Busta et al. adapted the YOLOv2 architecture and added a rotation parameter~\cite{Busta18}. They use bilinear sampling to rectify the word images and a direct application of CTC to do recognition. Shi et al. in~\cite{Shi18} introduce a full perspective rectification of words based on spatial transformer networks.

\noindent
In this work we do not apply a bounding box proposal approach and
instead detect text based on \emph{line proposals} derived from a
Hough Transform. The advantage of this technique is that we are not
required to generate multi-oriented bounding box proposals which
requires four coordinates for each proposal. Also, multi-oriented bounding
box proposals require complex post processing. 

\minisection{Descriptors for word recognition}  

\noindent
The proposal-based approaches discussed above work directly on image content, however
there are a few techniques that work on higher-level
representations. The Pyramidal Histogram Of Characters (PHOC) approach
for word spotting was proposed in~\cite{Almazan14}. The PHOC encodes
the spatial character distribution within words in a binary
vector. In~\cite{Bazazian18} a word spotting technique was proposed which recognizes characters individually and extracts a bounding box of each query word according to text proposals.
In comparison, we propose a model capable of learning characters and words
simultaneously. In addition, for the localization of the query words
we employ a novel line detection technique instead of text proposal
pipeline.

\minisection {Our contributions with respect to the state-of-the-art}

\noindent
In this work we extend the idea of the PHOC descriptor by directly
embedding at a pixel level the histograms of characters which can be
both synthesized from a given string or learned by a DCNN.  We employ a character recognition CNN while in~\cite{Li17} a text recognition network is applied. The key advantage of our approach is that we learn characters individually and independently of a lexicon of
words. In lieu of text bounding box proposals, we show how our
character probability maps can be used in a Hough Transform to propose
a compact set of candidate text lines for recognition.

\section{The Soft-PHOC Descriptor}
\label{sec:PropMethod}

We propose an FCN (Fully Convolutional Network \cite{Shelhamer16})
model to reproduce a special flavor of Soft-PHOC in order to learn
intermediate image representations that loosely correspond to
character heatmaps. By looking at words in a pyramidal way we are able
to capture spatial character dependencies and obtain a soft
probability distribution of characters over word patches in scene
images. Afterwards, we show how this mid-level representation of the
image can be used to tackle a word spotting task.

 \begin{figure*}
	\centering
	\includegraphics[width=0.85\textwidth, height=7cm]{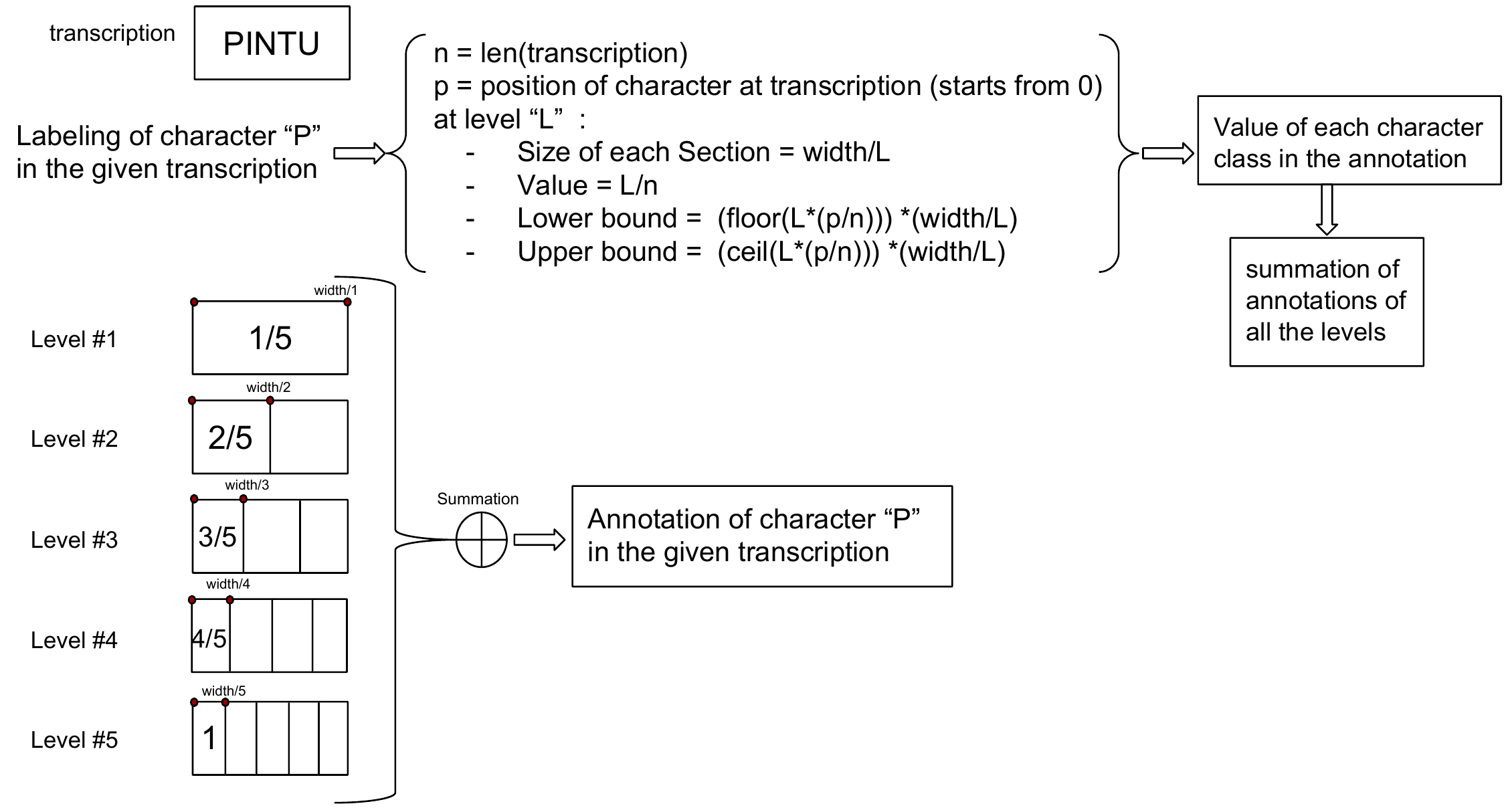} 
	\caption{Soft-PHOC annotation. For instance, if the transcription is
		``PINTU'', we show how we can define the annotation of class ``P''
		for the given transcription based on the value at each level of
		soft-PHOC descriptor.}
	\label{fig:annot_descriptor}
\end{figure*}

\subsection{Soft-PHOC labeling} 
\label{sec:PropMethod}

Exploiting the concept of PHOC (Pyramidal Histogram Of Characters)~\cite{Almazan14}, we devised a procedure for labeling text in natural scene images in order to learn correlations between characters within words. Each labeling is encoded directly in the image, establishing a correspondence between the pixels of the image and the elements of the annotation. We first present our labeling strategy by synthesizing the annotation from a word transcription and then we show how to encode it with reference to the spatial extent of the word in the scene.

As in standard PHOC, we follow a pyramidal approach to build the histogram of characters but we make the number of levels in the pyramid depend on the number of characters in the word $n$. Therefore, for each word we set the number of levels in the pyramidal representation equal to the number of characters it contains. For each level $L = {1,...,n}$ in the pyramid, we divide the annotation into $L$ spatial segments which we use as bins to build an histogram of characters. All the histograms for the different levels are then summed together to provide a compact and fixed-size labeling of the transcription, that encodes the position of its characters.
This is possible since every histogram is encoded as a fixed length patch with bins of varying width, depending on the level. The resulting annotation is a tensor of size $H\times W\times C$ where $H$ and $W$ can be arbitrarily chosen and $C$ is the number of class characters (38 including alphanumeric characters plus an additional one for punctuation). Since we want to encode the annotation inside the image reference frame, we choose $H$ and $W$ equal to the height and width of the rectified cropped word we are encoding.

Therefore, at each level $L$, the annotation tensor is divided into a corresponding number of bins, to which we assign the characters while building the histogram. Since the pixels occupied by the characters and by the bin sections may not be perfectly overlapped, we define the lower and upper bounds of the region of influence for the character at position $p$ as:
\begin{eqnarray}
l_p &=&  \mathrm{floor}(L*(p/n))*(W/L) \\
u_p &=&  \mathrm{ceil}(L*(p/n))*(W/L) 
\end{eqnarray}
where the $l_p$ and $u_p$ define the region of interest of the character within the word (i.e. the pixels corresponding to the bin of the histogram). The final annotations are then L1-normalized in order to obtain a valid character probability distribution for each pixel. The described annotation procedure is depicted in Fig.~\ref{fig:annot_descriptor}.

\begin{figure}[tbh]
	\centering
	\includegraphics[width=0.18\columnwidth]{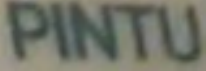}
	\includegraphics[width=\columnwidth]{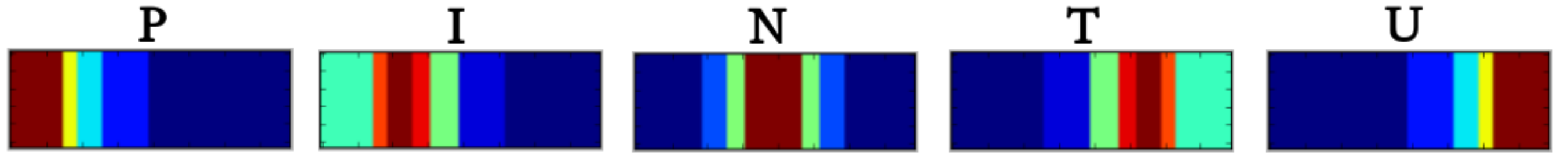}
	\caption{Example labeling for each character in cropped word
		images. The word in this example is ``PINTU''. First row shows the original cropped word and the second row shows the annotations. Each segment shows
		the heatmap of each character in the annotation. Note how the
		probability distribution of each character has spatial extent.}
	\label{fig:softPHOC}
\end{figure}

\begin{figure}[tbh]
	\centering
	\includegraphics[width=\columnwidth]{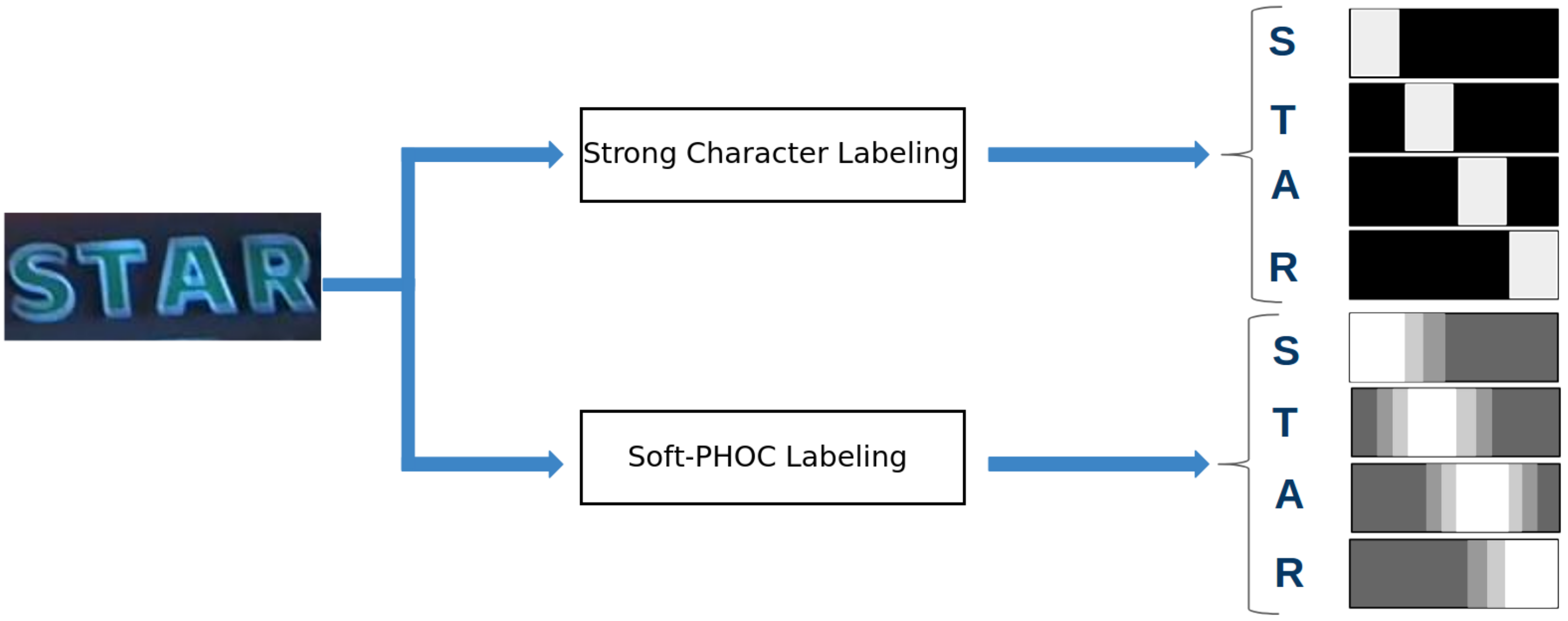} 
	\caption{Comparison of the annotation techniques. In this example the comparison is between strong character labeling technique such as~\cite{Bazazian18} and our proposed Soft-PHOC labeling. The darker color means lower value and the lighter color means higher value.} 
	\label{fig:softPHOC_strongcharatcter}
\end{figure}

\begin{figure*}
	\centering
	\includegraphics[width=0.7\textwidth]{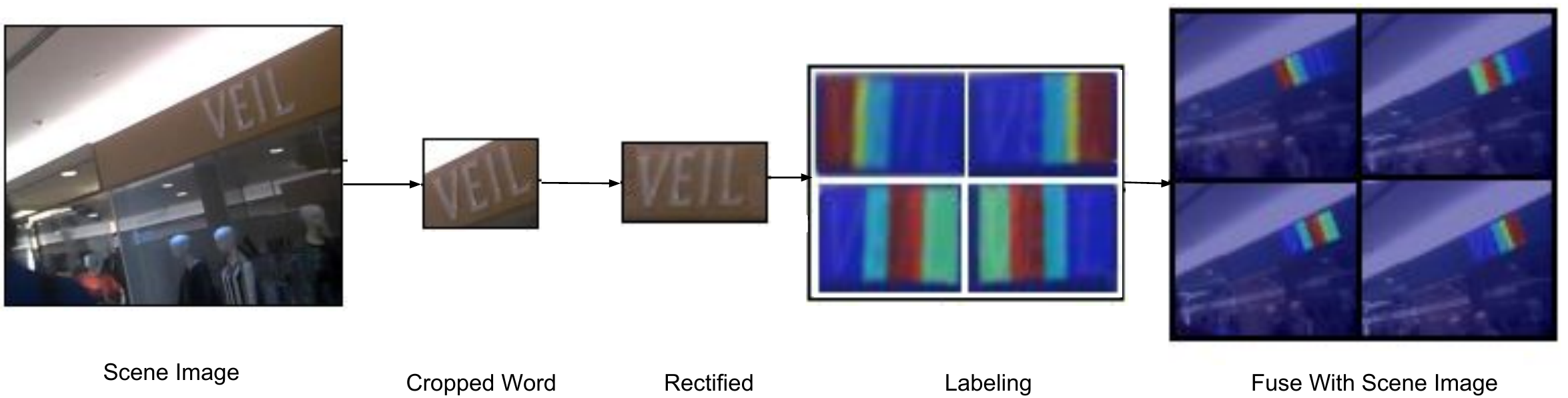} 
	\caption{Generating Soft-PHOC annotation in the scene images. First, the text region in the scene image should be cropped. Then, it should be rectified in order to define the Soft-PHOC annotation for the correspond transcription. Next, the character distributions interpolate  across the scene image based on its original localization and orientation. } 
	\label{fig:perspective}
\end{figure*}

The obtained annotation is a rectangular $C$-dimensional tensor with the same size of the rectified text crop. Each channel spatially encodes the probability of each pixel of belonging to a certain character. An example is given in Fig.~\ref{fig:softPHOC} where nonzero channels are shown. Afterwards, the annotation is projected back into the image to its original position (Fig.~\ref{fig:perspective}).
A comparison of the annotation techniques also illustrated in Fig.~\ref{fig:softPHOC_strongcharatcter}. In this example the comparison is between strong character labeling technique such as~\cite{Bazazian18} and our proposed Soft-PHOC labeling.

In order to extend this idea to scene images, we perform the steps shown in Fig.~\ref{fig:perspective} for each word in the scene. We start by cropping each text region and rectifying it to obtain an axes oriented rectangular patch. Afterwards, we define the Soft-PHOC embedding of the transcription, obtaining an arbitrary sized 38-dimensional tensor. The Soft-PHOC representations for each word are then fused together in a holistic representation of the scene by warping them back in the image reference system, maintaining their coordinates. The resulting annotation is a tensor with the same width and height of the image and a number of channels equal to the number of character classes we want to recognize (38).

\subsection{Soft-PHOCs Prediction}
Our annotation strategy is based on two motivations. First, given an image, we intend to be able to train an FCN~\cite{Shelhamer16} that produces the Soft-PHOC representation given an image. The FCN is supposed to output probability maps that can be interpreted locally as Soft-PHOC descriptors given a region in the image.
The training process does not depend on character-level annotations, although character-level annotations can be taken into account when available to provide a more precise labeling. Second, given the above annotation scheme, the network training phase is guided with a certain long-distance context about the existence of particular characters in different parts of the word. The concept is similar to the original Histogram of Characters idea on pyramidal levels~\cite{Almazan14}, but the information is encoded at pixel level instead of having a fixed length binary descriptor.

\subsection{Training Strategy}
\begin{figure*}
\centering
\includegraphics[width=\textwidth]{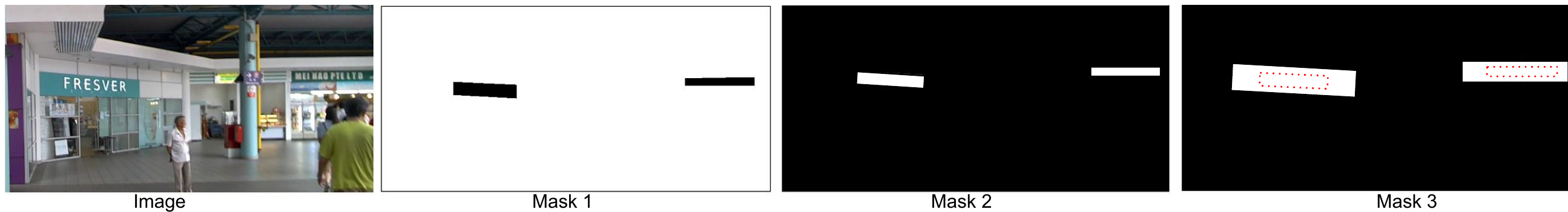} 
\caption{Three Masks for computing the loss: $Mask 1$ is for non-text
  pixels, $Mask 2$ for text pixels, and $Mask 3$ for text and a balanced
  subset of non-text pixels. To generate $Mask 3$ we add the half of
  width on the left and right side of each text region and half of
  height at the top and bottom. The dashed red line indicates the
  position of the ground-truth text box.}
\label{fig:Masks}
\end{figure*}

We use a network architecture inspired by FCN~\cite{Shelhamer16} to
learn to estimate Soft-PHOC labellings of scene images. In this way our network outputs an embedding of generic unlabeled images into the
Soft-PHOC space which can be compared with Soft-PHOC representations
of textual queries.  By construction, Soft-PHOC scene embeddings are
unbalanced since text areas are almost never predominant in the
scenes. In fact, we measured that scene text images contain around 90\%
background pixels and only 10\% text. At training time
this yields to an optimization problem with unbalanced classes which
is hard to optimize since the model focuses more on learning the
background rather than the text. This phenomena is even more important
in the case of character recognition, where each character class has
just a few pixels in comparison to the pixels for the background. To
solve this problem and balance the classes during training, we rely on
three different loss functions based on pixel-wise masks (see
Fig.~\ref{fig:Masks}) and their corresponding annotations:
\begin{itemize}
\item $\mathcal{L}_1$: focuses only on background pixels using a
  binary softmax cross entropy loss to discriminate between the
  background and the sum of the 37 remaining character classes.  We
  use a binary mask ($Mask 1$ in Fig.~\ref{fig:Masks}) to consider only
  non-text regions in the logit output map. This loss helps to model
  the background and lower the false positive rate.
\item $\mathcal{L}_2$: focuses on text regions and also uses a binary
  softmax cross entropy loss. This loss is complementary to
  $\mathcal{L}_1$ and uses precisely the complement of the background
  mask used for $\mathcal{L}_1$ ($Mask 2$ in Fig.~\ref{fig:Masks}) to
  focus only on text regions and allow the model to learn to localize
  words.
\item $\mathcal{L}_3$: focuses on the text regions plus a small
  background context around them by expanding the region of interest
  by 50\% in all directions ($Mask 3$ in Fig.~\ref{fig:Masks}). Again we use a softmax cross entropy
  loss, but this time over all the 38 classes (characters +
  background).
\end{itemize}

\begin{figure*}
\centering
\includegraphics[width=0.8\textwidth, height=7.5cm]{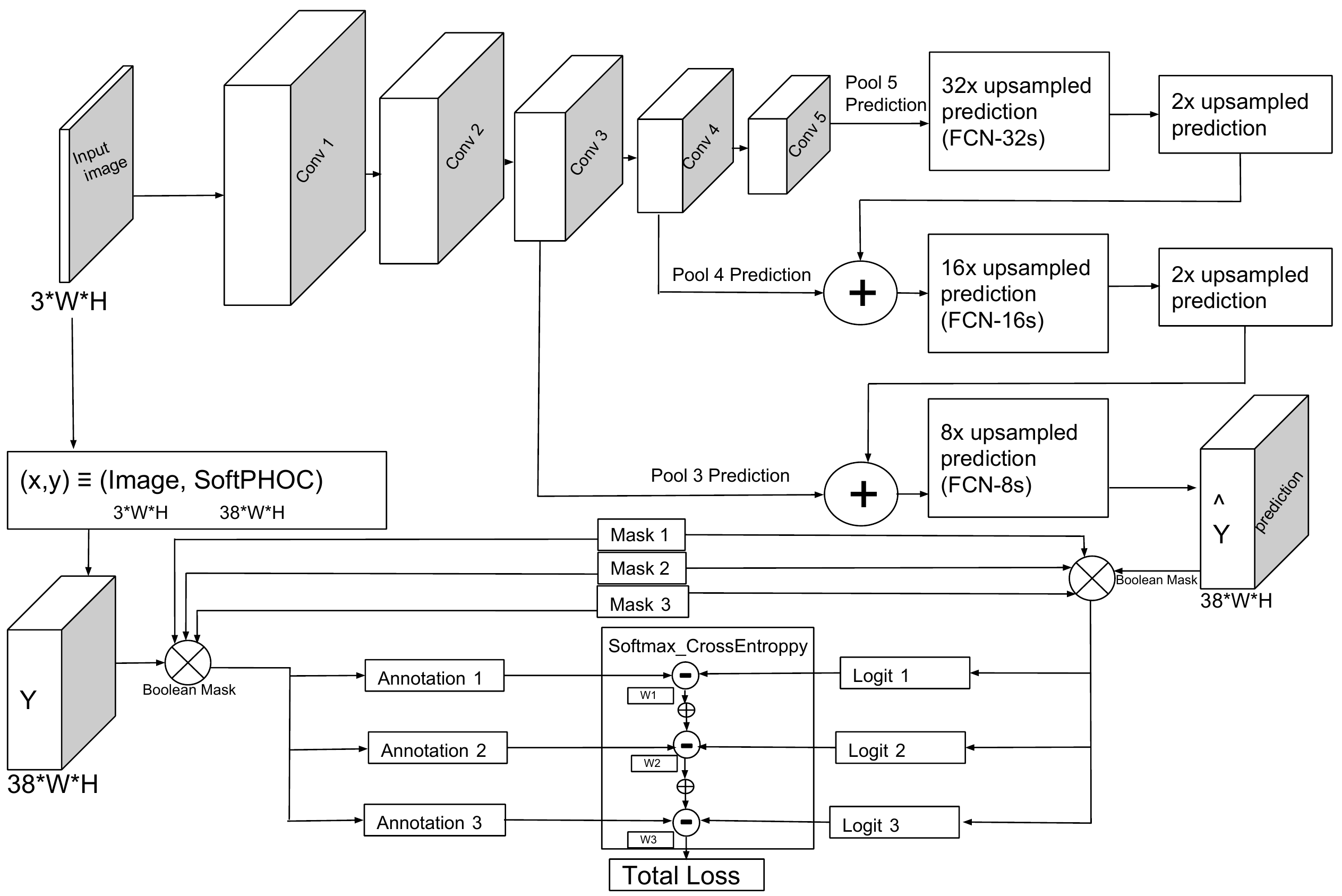} 
\caption{A Deep Convolutional Neural Network estimating Soft-PHOC
  descriptors. Our architecture is inspired by FCN~\cite{Shelhamer16}
  and its output is a pixel-wise character probability map over 38
  character classes at each spatial location in the
  image. Ground-truth Soft-PHOCs are synthetically generated from text
  annotations of word boxes over 38 character classes.}
\label{fig:net}
\end{figure*}

The three losses defined above are jointly optimized as a weighted sum loss:
\begin{eqnarray}
  \mathcal{L} &=& \alpha_1 \mathcal{L}_1 + \alpha_2 \mathcal{L}_2 + \alpha_3 \mathcal{L}_3.
\end{eqnarray}
In our experiments we use weights with increasing importance to make
the model focus on training to recognize characters more than the
background: $\alpha_1 = 0.1, \alpha_2 = 1, \alpha_3 = 2.5$.  We
trained our model using the Adam Optimizer with a learning rate of
$5\times10^{-5}$. The overall network architecture is shown in
Fig.~\ref{fig:net}. We have trained the network for 200K iterations based on the word-level annotations on the synthetic dataset~\cite{Gupta16}. Then, fin-tuned the model based on word-level annotations of ICDAR2015 challenge4-task4 dataset for 10K iterations.

\section{Line Detection for Queries}
\label{sec:HoughDTW}

In this section we explain the process of detecting a line for a given
word query. Each step of this process is explained in the following
sections as also shown in Fig.~\ref{fig:dtw}.

\subsection{Bigram Heatmap}
To obtain a query-specific heatmap, we use a the bigram approach. We consider the pixel-wise probability channels from the trained model that correspond to the characters that appear in the query word.
Moreover, we take into account the order of the characters in each
query beside its individual characters. The motivation of employing
bigrams is to distinguish between words with similar transcriptions,
for example anagrams such as ``listen'' and ``silent'', based on the order of the characters in each word.
The bigram probability $P_b$ of a given query is computed by
multiplying the character heatmaps two by two following the
transcription order as:
\begin{equation}
P_b = \sum_{i=1}^{n-1}(P(C_i)*P(C_{i-1})),
\end{equation}
where $n$ is the length of the query word (number of characters in the
query word), and $P(C_i)$ is the probability of the $i$th character in
the query.  For example, if the given query word is
``\textit{text}'', the bigram heatmp will be computed as:
$ (P(t)*P(e))+(P(e)*P(x))+(P(x)*P(t)).$

\subsection{Hough Transform}
Once a bigram heatmap for each given query is obtained, we threshold
it to generate a binary mask and apply Hough Transform on it. We use a
soft threshold by discarding all pixels with probability lower than
0.2. By applying the Hough Transform voting process on the binary
masks we obtain a set of text lines candidate for a query word.

\subsection{Dynamic Time Warping (DTW)}
In order to measure the similarity between each line in the proposal
set and the query we extract a Soft-PHOC representation of the line
and compare it with the descriptor of the transcription. To obtain the
Soft-PHOC of a line, we extract the corresponding pixels from the
38-dimensional output tensor of the network. Since each line $l$ has a
different length $L_l$, we obtain for each line a Soft-PHOC of size
$L_l\times1\times38$. On the other hand, for the textual query we
build a representation of size $n\times1\times38$, where $n$ depends
on the number of characters in the word.

\begin{figure*} 
\centering
\includegraphics[width=\textwidth]{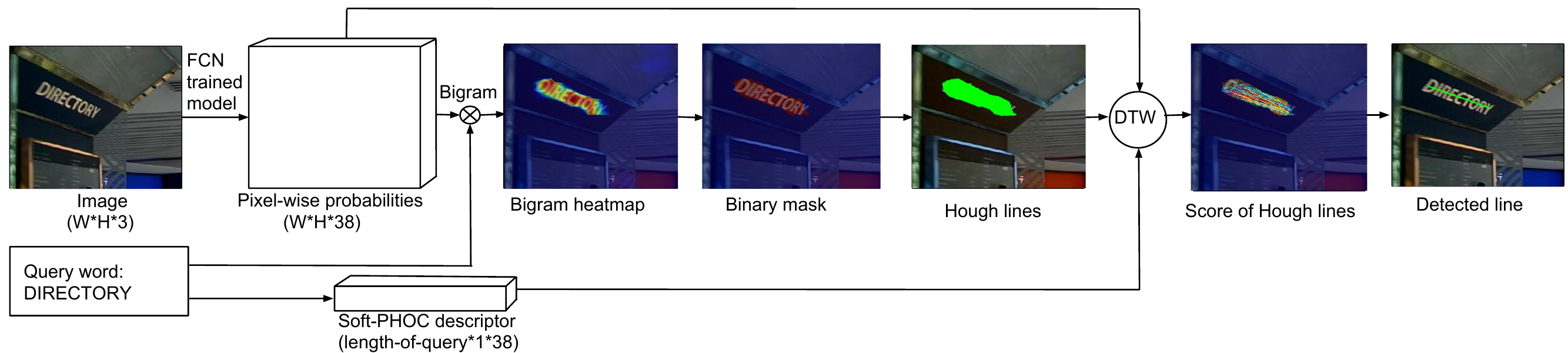} 
\caption{Line detection for each query. As input we have pixel-wise
  probabilities of each of the 38 character classes in the input
  image, plus a query word which in this example is ``DIRECTORY''. We
  generate a bigram heatmap for the corresponding query according to
  pixel-wise probabilistic from the net. Then, through a soft thresholding
  we create a binary mask that we use as input for the Hough Transform. From the
  stack of candidate Hough lines, we compute the distance between the
  Soft-PHOC from pixel-wise probabilities of each line and the
  Soft-PHOC descriptor of the query word. Since the text box width is
  variable at recognition time, we generate the query Soft-PHOC at a
  fixed width and use DTW to calculate the similarity with
  the estimated Soft-PHOC in the image.}
\label{fig:dtw}
\end{figure*}

To compare the line descriptors with the transcription, we use Dynamic
Time Warping to compute the similarities and find the line that best
suits the query. An example of the Hough lines that we obtain are
shown in Fig.~\ref{fig:dtw}, color coded with the similarity scores
provided by DTW.


\section{Experimental Results}
\label{sec:ExpResults}

We have evaluated the proposed Soft-PHOC descriptor in a word spotting application. The goal of word spotting is to localize in each image a list of given query words. The benchmark of the experiments is ICDAR2015-Challenge4, which are incidental images taken from a wearable ego-vision device. In addition, this dataset provides Strongly Contextualized list of query words which consists of 100 words per each image, including all the words that appear in the image and a number of distractor words. 

We have considered two evaluation techniques. One is based on detecting a text line for each query and the other is based on bounding box detection. In both cases, as shown in Fig.~\ref{fig:houghLines}, we take the output of the model and combine the channels correspondent to the query word characters, in order to obtain heatmaps for the bigrams that compose that word. By accumulating and thresholding the heatmaps we get a binary mask for each query. This allows us to have a region of attention specific to each query, which helps to focus only on relevant regions of the scene. Then, by applying the Hough Transform technique, we find a set of proposal lines for each query.

In order to recognize the correct location of a word we crop the whole output map in correspondence of each proposal line we sample the output map along each proposal line.
and compare it to the Soft-PHOC embedding of the query through Dynamic Time Warping (DTW). Based on the distance scores from DTW, we find the best candidate line of each query.  

In order to evaluate the detected line for each query we propose two methods. We describe our experimental results based on the line detection of queries in~\ref{subsec:softPHOC_eval}. Also, since the research community tends to represent words based on bounding boxes instead of line segments as proposed here, in order to compare our method with the state-of-the-art in word spotting, we applied a bounding box extraction technique which is explained in~\ref{subse:SOA_comparing}.
It has to be noted that unlike other methods that localize and then recognize, in our case the query is driving the localization process. The resulting line segments and bounding boxes that we obtain in the two evaluations are therefore guided by the query itself, and not by cues common to text in general.

\subsection{Soft-PHOC Evaluation}
\label{subsec:softPHOC_eval}


In the first evaluation technique, we compare the overlap of each line with its correspondent ground-truth bounding box as shown in Fig.~\ref{fig:houghLines} d(1) and e(1). We consider the line as a correct match when the line-box overlap measure exceeds a threshold $T$. We evaluate the results for varying thresholds $T=[0.3, 0.5, 0.7]$ and compute Precision, Recall and Accuracy. Results are reported in Tab.~\ref{tab:IoU_Threshold}. 

\subsection{Comparing Soft-PHOC with State-of-the-art}
\label{subse:SOA_comparing}
In the second evaluation method we produce a bounding box starting from the detected text line of each query. If the angle of the detected line segment is between $\pm45^{\circ}$, we define a vertical line (i.e. with an angle of $90^{\circ}$) passing in the middle of the horizontal line and with length equal to the one of the horizontal segment divided by number of characters in the query.
Accordingly, if the angle of the detected line is between $90^{\circ}\pm45^{\circ}$, we define a horizontal line (i.e. with an angle of $0^{\circ}$) which passes in the middle of the detected line and with length equal to the one of the vertical segment multiplied by the number of characters in the query word.
Therefore, we can obtain a final bounding box by considering these two lines as its axes.

\begin{figure*}
\centering
\includegraphics[width=\textwidth]{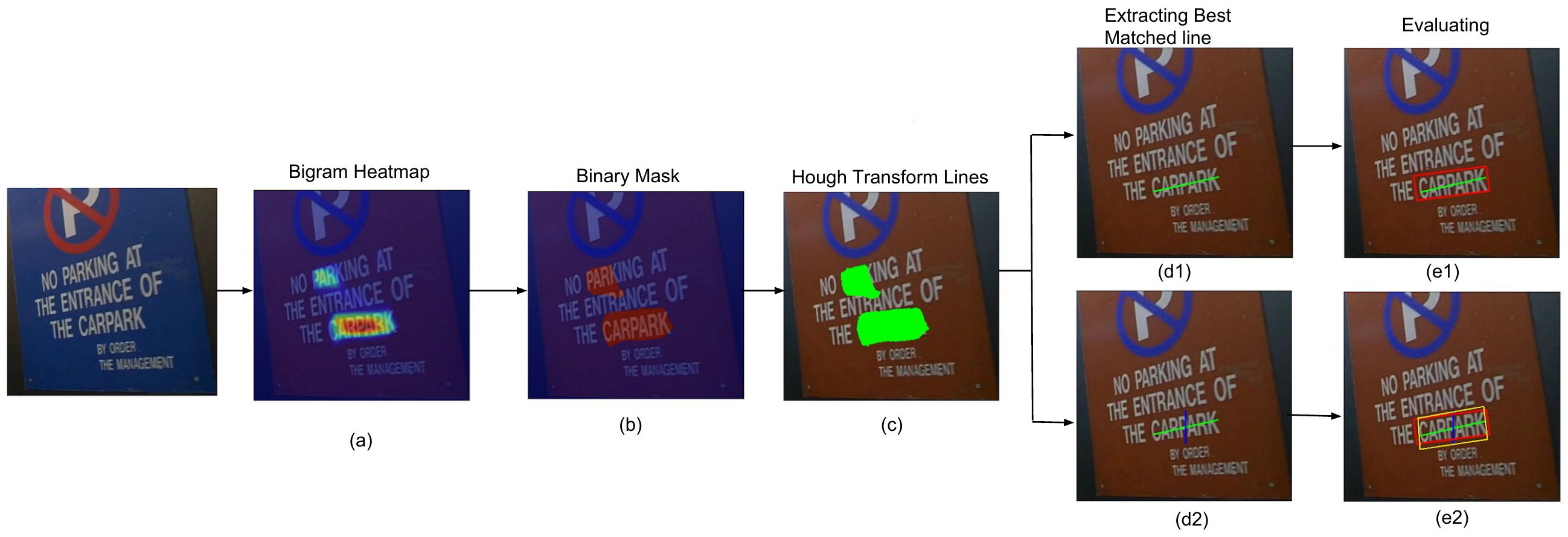} 
\caption{Extracting a sample query (CARPARK) at a scene text image. At (a) we have a bigram heatmap of the query. (b) a binary mask. (c) stack of detected hough lines from the binary mask. (d1) find the best matched line (green) of the query. (e1) evaluate the line with the ground-truth bounding box (red). (d2) define a vertical line (blue) for the best matched line (green) and (e2) extract a bounding box (yellow) from these two detected lines and compare it with the ground-truth bounding box (red).  }
\label{fig:houghLines}
\end{figure*}
\begin{figure*}
\centering
\includegraphics[width=0.85\textwidth, height= 7cm]{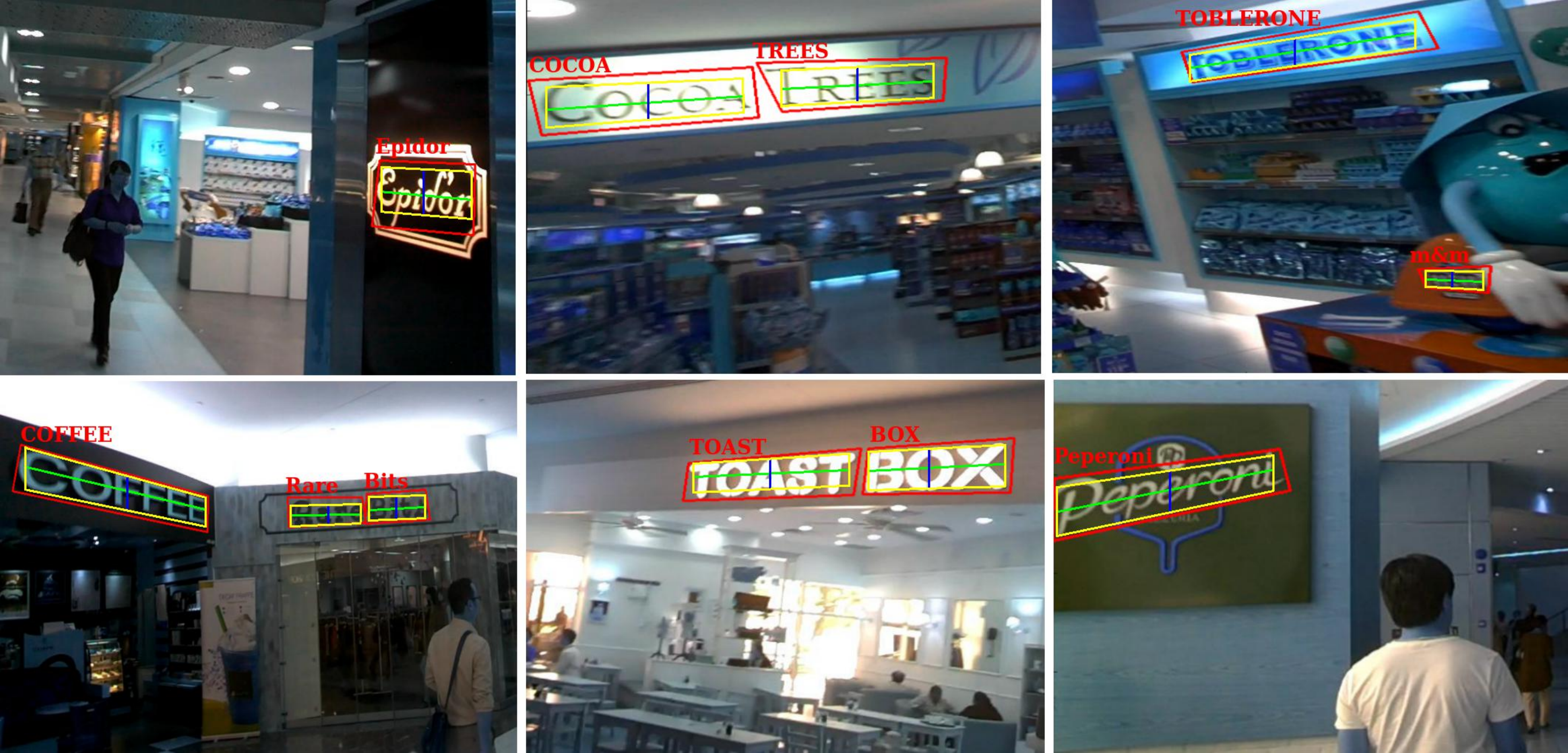} 
\caption{Qualitative results, the green lines are the detected lines from DTW and the blue line is the vertical line. Yellow bounding box is the extracted bounding box from the lines and red bounding box is the ground-truth.}
\label{fig:qualitative_results}
\end{figure*}
\begin{figure*}
\centering
\includegraphics[width =\textwidth]{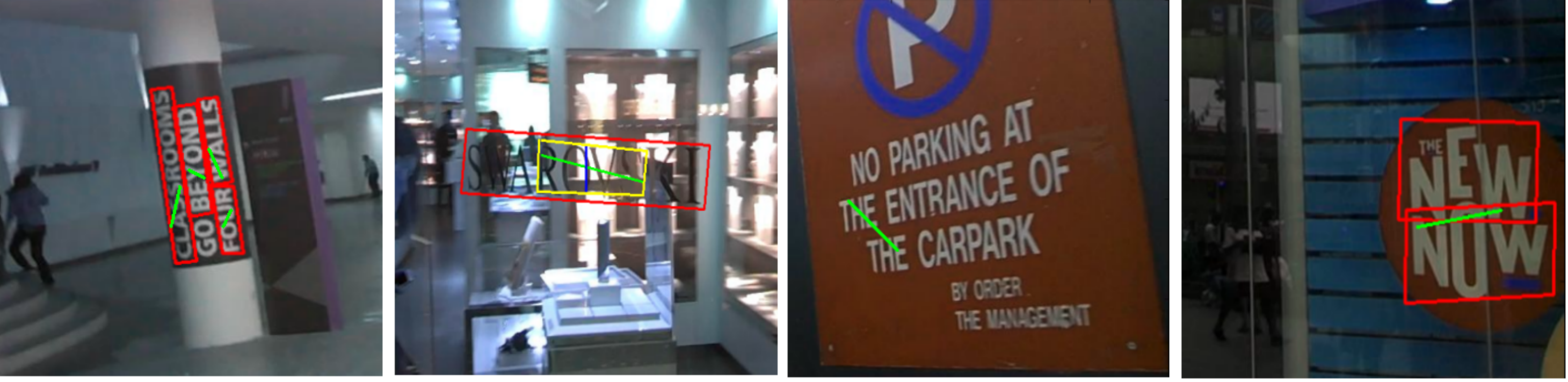} 
\caption{Failure cases. From left to right, the first image contains vertical text, the red bounding boxes are the ground-truth and the green lines are the best matched line for each query. The second image is showing the problem of incorrect localization due to clutter affecting the responses for some characters. The third image has two similar words close to each other in the same image. The green line shows the best matched line for query word of ``THE''. The fourth image is related to the cases when the word is written in an unconventional format and the detected line (in green for the word ``NOW'') results to have a wrong localization, in this case both words start and end with the same letter, so what you get is the average location between the two words. 
}
\label{fig:fail_exp}
\end{figure*}

Consequently, the resulting bounding box is compared with the ground truth location as shown in Fig.~\ref{fig:houghLines} d(2) and e(2). Quantitative results are as $Precision = 0.25$ and $Recall = 0.23$ and $Hmean = 0.24$
while qualitative results are shown in Fig.~\ref{fig:qualitative_results}. Failure cases are also reported in Fig.~\ref{fig:fail_exp}.

The main purpose of trying the second evaluation was based on two reasons: first, in order to demonstrate the capability of our proposed method for extracting bounding box in case if it required. Second, to be able to compare it with existing state-of-the art techniques since in all previous text spotting work the evaluation was based on bounding box detection.
According to the total numerical results of this section 
our results for word spotting seems to be far from the most recent state-of-the-art results such as~\cite{Liu18,Ma18,Shi18}. 
However, our proposed method is strong enough to detect accurately a rough spot of the localization of each query word even in the clutter background scene or images with the plenty of texts.

\begin{table}
  \centering
\caption{Evaluation of overlapping the selected line for each query and the ground-truth bounding box.}
\label{tab:IoU_Threshold}
\begin{tabular}{|c||c||c||c|}
  \hline
\textbf{Overlap Threshold} & \textbf{Precision}  & \textbf{Recall} & \textbf{Accuracy}  \\ \hline \hline
0.3 & 0.65 & 0.61 & 0.63 \\ \hline
0.5 & 0.58 & 0.57 & 0.58 \\ \hline
0.7 & 0.57 & 0.52 & 0.54 \\ \hline
\end{tabular}
\end{table}



\section{Conclusions and future work}
\label{sec:Conclusion}
In this paper we proposed a model to define an intermediate
representation of images based on text attributes which are individual
characters. Our representation, the Soft-PHOC, maintains local
information about the character distribution of entire words. To
perform word spotting we use a bigram heatmap of each query and build
a text line proposal algorithm based on the Hough Transform over the
character attribute maps of each query. Then, we detect the region of
each query by scoring the Hough lines using Dynamic Time Warping to
account for variable length word images. We evaluated our proposed
model in two different scenarios: one based on line detection and the
other on bounding box detection. Our preliminary experiments indicate that
detecting lines in comparison with a bounding boxes is geometrically
simpler and more efficient for detecting query words in scene
images.

We believe that detecting accurate bounding boxes is inconsequential for
the task of reading text. Defining lines just by two points makes
the detection process much more efficient and simpler due to the nature of text
in scene images which can have the various scales and orientations.

As future work we will extend the proposed model and fuse it with a
language model in order to perform open recognition in natural scene
images.


\section*{Acknowledgement}
\noindent 
This work has received funding from the project TIN2017-89779-P (AEI/FEDER, UE).


{\small
\bibliographystyle{ieee_fullname}
\bibliography{egbib}
}

\end{document}